\documentclass{article}

\usepackage{arxiv}

\usepackage{amsmath,amsfonts,bm}

\def\eqref#1{equation~\ref{#1}}

\def\1{\bm{1}}

\DeclareMathAlphabet{\mathsfit}{\encodingdefault}{\sfdefault}{m}{sl}
\SetMathAlphabet{\mathsfit}{bold}{\encodingdefault}{\sfdefault}{bx}{n}

\newcommand{\E}{\mathbb{E}}

\DeclareMathOperator*{\argmin}{arg\,min}

\usepackage{tikz}
\usetikzlibrary{arrows}
\usetikzlibrary{shapes}
\usepackage{multirow}
\usepackage{graphicx}
\usepackage{subcaption}

\usepackage{caption,setspace}
\newcommand{\fp}{{FP}\hspace*{2pt}}
\newcommand{\fn}{{FN}\hspace*{2pt}}

\newcommand{\pre}{{ Precision}\hspace*{2pt}}
\newcommand{\rec}{{ Recall}\hspace*{2pt}}
\newcommand{\fm}{{ F-measure}\hspace*{2pt}}

\newcommand{\remove}[1]{}

\newif\ifshowcomment
\ifshowcomment
    \newcommand{\todo}[1]{\textsf{\color{red}{[{TODO: #1}]}}}
    \newcommand{\mind}[1]{\textsf{\color{cyan}{[{Min: #1}]}}}
    \newcommand{\highlight}[1]{{\color{blue}{#1}}}
    \newcommand{\dawn}[1]{\textsf{\color{purple}{[{Dawn: #1}]}}}
\else
    \newcommand{\todo}[1]{}
    \newcommand{\mind}[1]{}
    \newcommand{\highlight}[1]{{\color{black}{#1}}}
    \newcommand{\dawn}[1]{}
\fi

    \newcommand{\revise}{\textcolor{black}}

\usepackage{thmtools,thm-restate}

\newtheorem{Df}{Definition}

\newcommand{\A}{\mathcal{A}}
\newcommand{\D}{\mathcal{D}}

\usepackage{thmtools}
\usepackage{thm-restate}

\usepackage{hyperref}

\usepackage{cleveref}

\usepackage[utf8]{inputenc} 
\usepackage[T1]{fontenc}    
\usepackage{hyperref}       
\usepackage{url}            
\usepackage{booktabs}       
\usepackage{amsfonts}       
\usepackage{nicefrac}       
\usepackage{microtype}      
\usepackage{lipsum}

\title{robust anomaly detection and backdoor attack detection via differential privacy}

\author{
Min Du, Ruoxi Jia, Dawn Song \\
University of California, Berkeley \\
\texttt{\{min.du,ruoxijia,dawnsong\}@berkeley.edu} 
}

\begin{document}
\maketitle

\begin{abstract}

Outlier detection and novelty detection are two important topics for anomaly detection. Suppose the majority of a dataset are drawn from a certain distribution, outlier detection and novelty detection both aim to detect data samples that do not fit the distribution. \textit{Outliers} refer to data samples within this dataset, while \textit{novelties} refer to new samples. In the meantime, backdoor poisoning attacks for machine learning models are achieved through injecting poisoning samples into the training dataset, which could be regarded as ``outliers'' that are intentionally added by attackers. Differential privacy has been proposed to avoid leaking any individual's information, when aggregated analysis is performed on a given dataset. It is typically achieved by adding random noise, either directly to the input dataset, or to intermediate results of the aggregation mechanism. In this paper, we demonstrate that applying differential privacy can improve the utility of outlier detection and novelty detection, with an extension to detect poisoning samples in backdoor attacks. We first present a theoretical analysis on how differential privacy helps with the detection, and then conduct extensive experiments to validate the effectiveness of differential privacy in improving outlier detection, novelty detection, and backdoor attack detection.
\end{abstract}

\section{Introduction}
Given a dataset where most of the samples are from a certain distribution, outlier detection aims to detect the minorities in the dataset that are far from the distribution, while the goal of novelty detection is to detect newly observed data samples that do not fit the distribution.
\highlight{On the other hand, poisoning examples that are intentionally added by attackers to achieve backdoor attacks could be treated as one type of ``outliers'' in the training dataset.}
Using machine learning for outlier/novelty detection is typically to train a model that learns the distribution where the training data samples are drawn from, and the final trained model could give a high anomaly score for the outliers/novelties that deviate from the same distribution.
In both cases, the machine learning model is not supposed to learn from the outliers in the training dataset. Unfortunately, deep learning models that contain millions of parameters tend to remember too much (\cite{song2017machine}), and can easily overfit to rare training samples (\cite{carlini2018secret}).

Protecting data privacy has been a major concern in many applications, because sensitive data are being collected and analyzed.
Differential privacy has been proposed to ``hide'' certain input data from the output; that is, by looking at the statistical results calculated from input data, one cannot tell whether the input data contain a certain record or not.
The way of applying differential privacy is to add random noise to the input data or the data analysis procedure, such that the output difference caused by the input difference can be hidden by the noise. \revise{A known fact is that differential privacy implies stability (\cite{kasiviswanathan2011can}). Particularly, a differentially private learning algorithm is stable in the sense that the model learned by the algorithm is insensitive to the removal or replacement of an arbitrary point in the training dataset~\cite{bousquet2002stability}. When the training dataset contains a handful of outliers, the output model of a stable learning algorithm should be close to the one trained on the clean portion of the training set. Intuitively, compared with the model trained on contaminated dataset, the one trained on clean data could be better at distinguishing outliers from normal data. Therefore, differential privacy can potentially be leveraged to improve the identification of outliers. This motivates us to apply differential privacy to anomaly detection and defense against backdoor attacks.}
\remove{
It has been proposed that
differential privacy has the potential to prevent overfitting~\cite{pnml}, because of its ability to generalize from data~\cite{nissim2015generalization}, without depending on any single record.

Inspired by the ability to prevent overfitting, we apply differential privacy to outlier detection and novelty detection, to avoid learning from outliers, which shows significant utility improvement.
}


\paragraph{Our contribution.}
First, we present a theoretical explanation on why differential privacy can help to detect outliers from a training and \revise{testing} dataset, as well as an analysis on the relationship between the number of outliers to detect and the amount of random noise to apply.
Second, to demonstrate the effectiveness, we apply differential privacy to an autoencoder network trained on a constructed MNIST dataset with injected outliers, for both outlier detection and novelty detection, to show how much the utility could be improved with different amount of outliers and noise.
Third, we apply differential privacy to a real-world task - Hadoop file system log anomaly detection. 
\revise{System log anomaly detection is an important topic in computer security. Our proposed method greatly improves upon the state-of-the-art system in this field.}
The results 
indicate that differential privacy is able to eliminate almost all the false negatives, and achieve significantly improved overall utility, compared with the current state-of-the-art work DeepLog (\cite{du2017deeplog}).
\highlight{Finally, via a proof-of-concept experiment using MNIST dataset with injected poisoning samples, we show that the idea of outlier detection could be extended to backdoor attack detection, and that differential privacy is able to further improve the performance.}

\section{Preliminary}
\label{sec:dp_pre}
Given an input dataset and an aggregation mechanism, {\em differential privacy} (\cite{dwork2011differential}) aims to output the requested aggregation results, which are guaranteed not to reveal the participation of any individual data record.
Formally, differential privacy is defined as below:

\begin{Df}[Differential privacy]
A randomized mechanism $\mathcal{M}$ applied to data domain $\mathbb{D}$ is said to be $(\epsilon$, $\delta)$-differentially private if for any adjacent datasets $d$, $d'$ in $\mathbb{D}$, and any subset of outputs $\mathcal{S}\subseteq \text{Range}(\mathcal{M})$, it holds that
$$
\mathrm{Pr}[\mathcal{M}(d) \in \mathcal{S}] \le e^{\epsilon} \mathrm{Pr}[\mathcal{M}(d') \in \mathcal{S}] + \delta,
$$
\end{Df} 
where $\epsilon$ stands for the privacy bound, and $\delta$ stands for the probability to break this bound.

The \textit{adjacent} datasets $d$, $d'$ could be understood as two databases, where only one record differs, i.e., $\|d-d'\|_1=1$. Differential privacy guarantees that the difference between $d$ and $d'$ are not revealed through inspecting the outputs $\mathcal{M}(d)$ and $\mathcal{M}(d')$.
Clearly, the closer $\epsilon$ is to $0$, the more indistinguishable $\mathcal{M}(d)$ and $\mathcal{M}(d')$ are, and hence the stronger the privacy guarantee is. 

A common approach to enforcing differential privacy for a function $f: \mathbb{D} \rightarrow \mathbb{R}$, is to add random Gaussian noise $\mathcal{N}(0, \sigma ^ 2)$ to 
perturb the output in $\mathbb{R}$.
The intuition is that, for given adjacent datasets $d$ and $d'$, one cannot tell whether the difference between $f(d)$ and $f(d')$ is incurred by the single record that differs in $d$ and $d'$, or by the random noise being applied. \revise{The magnitude of Gaussian noise needs to be tailored to the maximum difference between $f(d)$ and $f(d')$, which is formally defined as $\mathcal{L}_2$-sensitivity.} 
\begin{Df}[$\mathcal{L}_2$-sensitivity]
The $\mathcal{L}_2$-sensitivity for a function $f: \mathbb{D} \rightarrow \mathbb{R}$ is: 
$$\Delta = \max\limits_{\substack{d,d'\in\mathbb{D} \\ {\|d-d'\|_1=1}}} \|f(d)-f(d')\|_2$$
\end{Df}
The 
noise scale $\sigma$ to apply
can be calculated as below (\cite{dwork2014algorithmic}).
\begin{restatable}{Thm}{goldbach}
\label{theo:sigma}
To perturb a function with sensitivity $\Delta$ under $(\epsilon, \delta)$ - differential privacy, the minimum noise scale $\sigma$ of Gaussian mechanism is given by: $\sigma = \frac{\Delta}{\epsilon} \cdot \sqrt{2 \ln \frac{1.25}{\delta}}$, where $\epsilon \in (0, 1)$.
\end{restatable}


\paragraph{Deep learning with differential privacy (\cite{abadi2016deep})}
The procedure of deep learning model training is to minimize the output of a loss function, through numerous stochastic gradient descent (SGD) steps. 
\cite{abadi2016deep} proposed a differentially private SGD algorithm that works as follows. At each SGD step, a fixed number of randomly selected training samples are used as a mini batch. 
For each mini batch training, the following two operations are performed to enforce differential privacy: 1) clip the norm of the gradient for each example, with a clipping bound $C$, to limit the sensitivity of \revise{gradient};
2) sum the clipped per-example gradients and add Gaussian noise $\mathcal{N}(0, \sigma^2)$, before updating the model parameters.
\revise{\cite{abadi2016deep} further proposed a moment accounting mechanism which calculates the aggregate privacy bound when performing SGD for multiple steps. Differential privacy is immune to post-processing. Therefore, the output of the trained model for any queries enjoys the same privacy guarantee as the above SGD-based training process.}

\section{The connection between differential privacy and outlier detection}
\label{sec:theory}
By definition, random noise added into model training for differential privacy hides the influence of a single record on the learned model. Intuitively, if applying differential privacy to the training process, the contribution of rare training examples will be hidden by random noise, resulting in a model that underfits the outliers. Such model will facilitate novelty \revise{and outlier} detection because it will be less confident in predicting the atypical examples. In this section, we first present a theorem to precisely characterize the above intuition, and then analyze the relationship between the number of outliers in the training dataset and the amount of noise to apply.

\paragraph{Notations} Let $\mathcal{Z}$ be the sample space and $\mathcal{H}$ be the hypothesis space. The loss function $l:\mathcal{H}\times\mathcal{Z}\rightarrow\mathbb{R}$ measures how well the hypothesis $h\in\mathcal{H}$ explains a data instance $z\in\mathcal{Z}$. A learning algorithm $\mathcal{A}:\mathcal{Z}^n\rightarrow \mathcal{H}$ learns some hypothesis $\mathcal{A}(S)$ given a set $S$ of $n$ samples. For instance, in supervised learning problems, $\mathcal{Z} = \mathcal{X}\times \mathcal{Y}$, where $\mathcal{X}$ is the feature space and $\mathcal{Y}$ is the label space; $\mathcal{H}$ is a collection of models $h:\mathcal{X}\rightarrow\mathcal{Y}$; and $l(h,z)$ measures how well $h$ predicts the feature-label relationship $z=(x,y)$. 

Let $S=\{z_1,\ldots,z_n\}$ be a set of independent samples drawn from an unknown distribution $\mathcal{D}$ on $\mathcal{Z}$. For a given distribution $\mathcal{D}$, 
an \emph{oracle} hypothesis 
is
the one that minimizes the expected loss:
\begin{align}
    h^* = \argmin_h \mathbb{E}_{z\sim \mathcal{D}}[l(h,z)]
\end{align}

We define an outlier as a data instance that has significantly different loss from the population under the oracle hypothesis.

\begin{Df}
We say $\tilde{z}$ is an outlier with regard to distribution $\mathcal{D}$ if 
\begin{align}
   l(h^*,\tilde{z}) -  \mathbb{E}_{z\sim \mathcal{D}} [l(h^*,z)] \geq T
\end{align}
where $T$ is a constant that depends only on $\mathcal{D}$.
\end{Df}

We will prove the usefulness of differential privacy to detect outliers for the classes of learning algorithms that produce hypotheses converging to the optimal hypothesis asymptotically pointwise. We define such learning algorithms to be \emph{uniformly asymptotic empirical risk minimization} (UAERM).

\begin{Df}
\label{def:uaerm}
A (possibly randomized) learning algorithm $\mathcal{A}$ is UAERM with rate $\xi(n,\mathcal{A})$ if for any distribution $\mathcal{D}$ defined on the domain $\mathcal{Z}$, it holds that
\begin{align}
   \forall z \quad \quad |\mathbb{E}_{\mathcal{S}\sim \mathcal{D}^n} \mathbb{E}_{h \sim \mathcal{A}(S)} l(h,z) - l(h^*,z)| \leq \xi(n,\mathcal{A})
\end{align}
\end{Df}

In the definition, we make it explicit that the rate $\xi(n,\mathcal{A})$ depends on the learning algorithm $\mathcal{A}$. For instance, if $\mathcal{A}$ is a differentially private learning algorithm, the rate will depend on the privacy parameters. In that case, with slight abuse of notation, we will denote the rate for a $(\epsilon,\delta)$-differentially private learning algorithm trained on $n$ data instances by $\xi(n,\epsilon,\delta)$.

Due to the nonconvexity of their loss functions, neural networks may not enjoy a useful, tight characterization of the learning rate. Thus, we will empirically verify that using noisy SGD to learn differentially private neural networks is UAERM. Moreover, as we will show in the experiment, $\xi(n,\epsilon,\delta)$ grows as privacy parameters $\epsilon$ and $\delta$ become smaller. Intuitively, this is because larger noise is required to ensure stronger privacy guarantees, which, on the other hand, slows down the convergence of the learning algorithm.

Without loss of generality, we assume that $0\leq l(h,z)\leq 1$. The following theorem exhibits how the prediction performance of differentially private models on normal data will differ from outliers and connects the difference to the privacy parameters of the learning algorithm and the amount of outliers in the training data.

\begin{restatable}{Thm}{goldbach}
\label{thm:gap}
Suppose that a learning algorithm $\mathcal{A}$ is $(\epsilon,\delta)$-differentially private and UAERM with the rate $\xi(n,\epsilon,\delta)$. Let $S' = S\cup U$, where $S\sim \D^n$ and $U$ contains $c$ arbitrary outliers. Then 
\begin{align}
\label{eqn:bound}
    &\E_{h\sim \A(S')} l(h,\tilde{z}) - \E_{h\sim \A(S')} \E_{z\sim \D}l(h,z)  \nonumber\\
    &\quad\quad\geq T -  2\bigg(\xi(n,\epsilon,\delta) + \sqrt{\frac{n(e^{\epsilon}-1+\delta)^2}{2}\log \frac{2}{\gamma}} + e^{c\epsilon} - 1 + ce^{c\epsilon}\delta\bigg)
\end{align}
with probability at least $1-\gamma$.
\end{restatable}


The two terms in the left-hand side of (\ref{eqn:bound}) represent the model's prediction loss on outliers and normal test data drawn from $\D$, respectively. Due to the stochasticity of differentially private learning algorithms, the difference is characterized by the expectation taken over the randomness of the learned models. The theorem establishes a lower bound on the prediction performance difference between normal and outlier data. A larger difference indicates that identifying outliers will be easier. 

The impact of privacy parameters on the lower bound manifests itself in two aspects. On one hand, stronger privacy guarantees (i.e., smaller $\epsilon$ and $\delta$) will require higher noise to be added into the training process, which increases the learning rate $\xi(n,\epsilon,\delta)$. On the other hand, increasing privacy level will improve the stability of the learning algorithm; the resulting models tend to ignore the outliers in the training set and become closer to the ones trained on completely clean data, thus making the outlier detection more effective. The second aspect is embodied by the fact that the terms except $\xi(n,\epsilon,\delta)$ in the parenthesis of the lower bound grow with $\epsilon$ and $\delta$. Therefore, the privacy parameters cannot be too large or too small in order to ensure optimal anomaly detection performance. 

Moreover, the relationship between the right-hand side of (\ref{eqn:bound}) and $c$ indicates that the anomaly detection problem will be more difficult with more outliers in the training set. Dissecting the right-hand side of (\ref{eqn:bound}), we further observe that $c$ appears always in tandem with $\epsilon$. This implies that for larger number of outliers in the training set (i.e., $c$ is larger), we will need to tune down $\epsilon$ and $\delta$ to maintain the same novelty detection performance. 

\revise{Last but not least, the definition of outliers in our paper is quite general---it does not make any assumptions about how the outliers are generated. Also, we do not make assumptions about whether these outliers are in training or test data. Therefore, our analysis can shed light on detecting various types of anomalies, including but not limited to outlier/novelty detection, backdoor detection, and noisy label detection. In the following experimental section, we will focus our evaluation on outlier/novelty detection and defense against backdoor attacks. }

\section{Experiments}
\label{sec:exp}
This section empirically evaluates the effectiveness of differential privacy in improving anomaly detection and backdoor attack detection.
We call an outlier/novelty or a poisoning example 
as a \textit{positive}, and other normal data samples as \textit{negatives}.
The metrics measured by each experiment include:
false positive (FP), false negative (FN), \pre = TP / (TP+FP), \rec= TP / (TP+FN),
Area under the receiver operating characteristic curve (AUROC) which is the area under the TPR-FPR curve, Area under the \pre-\rec curve (AUPR) which summarizes the \pre-\rec curve, as well as \fm=2$\times$ \pre $\times$ \rec / (\pre $+$ \rec). The detailed explanations of all these measures could be found in \cite{wiki:fpfn,wiki:fm,ap,auc}.


\subsection{Outlier detection and novelty detection with autoencoders}
\label{sec:exp-auto}
Autoencoder is a type of neural network that has been widely used for outlier detection and novelty detection. It contains an encoder network which reduces the dimension of the input data, and a decoder network which aims to reconstruct the input. Hence, the learning goal of autoencoders is to minimize the 
reconstruction error,
which is consequently the loss function. Because the dimensionality reduction brings information loss, and the learning goal encourages to preserve the information that is common to most training samples, outliers 
that contain rare information
could be identified
by measuring model loss. 
In this section, with a varying amount of outliers and noise scale, we show how differential privacy would improve the utility of anomaly detection with autoencoders.

\paragraph{Datasets.} We utilize MNIST dataset composed by handwritten digits 0-9, and notMNIST dataset (\cite{notmnist}), which contains letters A-J with different fonts. 
The original MNIST data contain 
$60,000$ training images, and 
$10,000$ test images, which we refer to as \textit{MNIST-train} and \textit{MNIST-test} respectively.
The notMNIST data contain $10,000$ training images and $1,000$ test images, denoted as \textit{notMNIST-train} and \textit{notMNIST-test}.
Based on these datasets, we intentionally construct training datasets with varying amount of injected outliers.
Specifically, each training dataset is constructed with a particular outlier ratio $r_o$, such that the resulted dataset MNIST-OD-train($r_o$) contains $60,000$ images in total, where a percentage of $1-r_o$ are from MNIST-train, 
and $r_o$ are from notMNIST-train. 
For each training dataset MNIST-OD-train($r_o$), a set of autoencoder models are trained with varying noise scale $\sigma$ applied 
for differential privacy. 
For an autoencoder model trained on dataset MNIST-OD-train($r_o$), outlier detection is thus to detect the $r_o\times 60,000$ outliers from MNIST-OD-train($r_o$). 
For novelty detection, we further construct a test dataset MNIST-ND-test, which is composed by the entire MNIST-test dataset and notMNIST-test dataset, a total of $11,000$ images. The goal of novelty detection is to identify the $1,000$ notMNIST-test images as novelties. 

\paragraph{Evaluation metrics.}
To check whether a data sample is an outlier/novelty using autoencoders, the standard practice is to set a loss threshold based on training statistics, and any sample having a loss above this threshold is an outlier/novelty. 
To measure the performance under different thresholds,
we use the AUPR score which is a threshold-independent metric.
Compared with other metrics such as the AUROC score, the AUPR score
is more informative when the positive/negative classes are highly unbalanced, e.g., 
for
outlier detection where the ratio of outliers is extremely low.
\highlight{
More experiment settings with both AUPR and AUROC metrics 
are
in appendix which present similar observations. 
}

\paragraph{Set-up.}
For autoencoders,
the encoder network contains 3 convolutional layers with max pooling,
while the decoder network contains 3 corresponding upsampling layers.
For differential privacy, we use a clipping bound $C=1$ and $\delta=10^{-5}$, and vary the noise scale $\sigma$ as in \cite{abadi2016deep}.
All models are trained with a learning rate of 0.15, a mini-batch size of 200 and for a total of 60 epochs.

\begin{figure}[htbp]
\centering
\begin{minipage}{0.375\textwidth}
\centering
	\centering{
		\includegraphics[width=\linewidth]{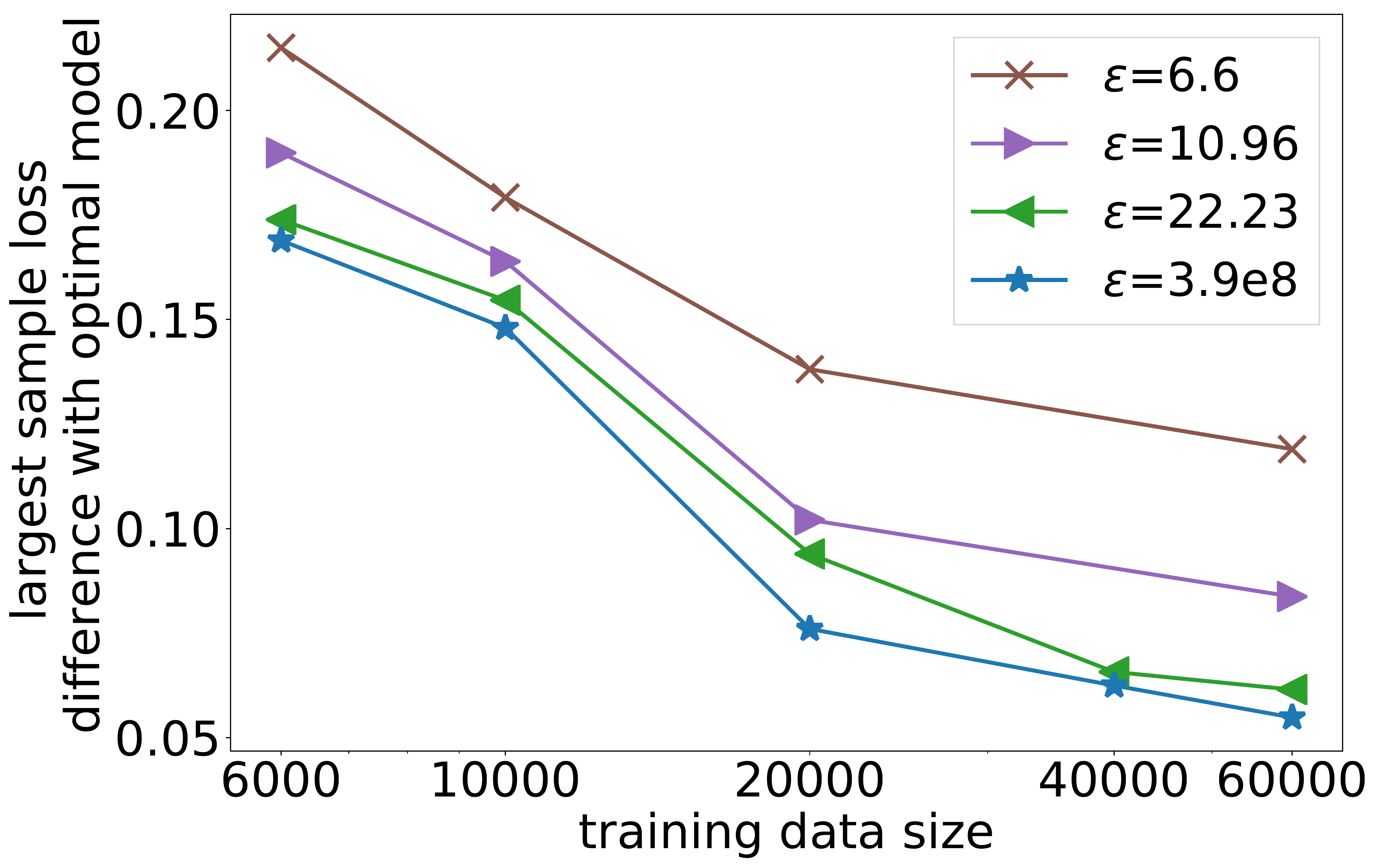}
	}
	\caption{The largest test sample loss between a differentially private model trained on a random subset of training data and the oracle hypothesis.}
	\label{fig:theo}
\end{minipage}
\hspace{+1mm}
\begin{minipage}{0.605\textwidth}
\centering
\captionsetup{type=table} 
\small
	\tabcolsep=0.1cm
    \begin{tabular}{ |l|r|r|r|r|r|r|r|r| } \hline
    noise   &  \multicolumn{8}{|c|}{outlier percentage in training data $r_o$}   \\  \cline{2-9}
    scale    & \multicolumn{2}{|c|}{$0.1\%$} & \multicolumn{2}{|c|}{$0.5 \%$} & 
    \multicolumn{2}{|c|}{$5 \%$} & \multicolumn{2}{|c|}{$10\%$}    \\ \cline{2-9}
    $\sigma=$ & OD & ND & OD & ND & OD & ND & OD & ND \\ \hline
     N/A  &  99.92 & 99.77 & 92.12 & 98.81 & 84.33 & 88.18 & 72.16 & 68.14     \\ \hline
    $\mathbf{0}$  &  
99.89 & 99.83  & 98.3& 99.69  & 83.86 & 87.91 & 77.8 & 74.74 
 \\ \hline
    $  0.01$  &  {\bf 100} & {\bf 99.97} &94.92 & 99.23  &90.79 & 93.34  &85.41 & 84.07
 \\ \hline


    $  0.1$  & 100 & 99.85 &
{\bf 98.44} & {\bf 99.66} &
92.23 & 94.21                     &
85.56 & 83.98
 \\ \hline

    $  1$  &  100 & 99.78 &98.28 & 99.67 &94.92 & 96.87 &81.87 & 80.12
\\ \hline

    $  5$  &  99.87 & 99.49 & 98.51 & 99.52 &
{\bf 96.78} & {\bf 98.04} &
95.25 & 95.41
\\ \hline
$ 10$  & 90.24 & 97.77 & 91.88 & 98.2  &96.6 & 98.2  & {\bf 97.07} & {\bf 97.46} 
\\ \hline
\multicolumn{9}{|l|}{{\sf \scriptsize Below $\sigma$ value is too big such that the model does not converge well.}} \\ \hline
$  50$  &  65.94 & 92.13 & 70.34 & 90.8 & 86.58 & 91.59 & 88.49 & 90.27
\\ \hline
    \end{tabular}
	\caption{{AUPR scores for outlier detection (OD) and novelty detection (ND). $\sigma=0$ indicates applying clipping bound only.} 
	}
	\label{table:train_ae}
\end{minipage}
\end{figure}

\highlight{
\paragraph{Validation of UAERM for Noisy SGD.}
To begin with, we first conduct experiments to empirically validate our assumption in Theorem~\ref{thm:gap}. \revise{While a rigorous verification of the assumption is intractable as it requires the knowledge of underlying data distribution and computing expected loss over randomness of both data distribution and differentially private algorithms, our experiments provide a sanity check of the assumption by replacing the expectation by the empirical average of a large number of data samples.} For this set of experiments, we only utilize MNIST data for training, while the test dataset contains all available MNIST and notMNIST test samples. The oracle hypothesis is trained on all available training data, while each differentially private model is trained with varying privacy level $\epsilon$, and training data size.
\revise{For a fixed training set and $\epsilon$, we perform training for multiple times to accommodate the randomness of differentially private training. Further, we train on multiple randomly selected training sets of the same size. We measure the loss of each resulting model on the test set, and calculate the average testing loss across different runs of differentially private training and different randomly selected training sets of the same size. We then compute the largest difference between the averaged test loss and the test loss associated with the oracle hypothesis.} 
The results are shown in Figure~\ref{fig:theo}. Each data point in the figure is an average of 9 differentially private models trained on 3 randomly sampled subsets of the training data, and 3 random training processes for each sampled subset.
As in Figure~\ref{fig:theo}, the larger the training data size, and the larger $\epsilon$ is, the closer of the randomized model to the oracle hypothesis, validating our assumption in Theorem~\ref{thm:gap} that noisy SGD is UAERM.}

\paragraph{Detection results} 
Table~\ref{table:train_ae} shows the outlier detection (OD) results on dataset MNIST-OD-train, as well as the novelty detection (ND) results on dataset MNIST-ND-test. 
OD mimics the unsupervised anomaly detection case. 
ND mimics the case where the autoencoder model is supposed to be trained on normal data, to detect unforeseen anomalies, 
while the training dataset is noisy. 
The first row where $\sigma=$N/A is for the baseline model without differential privacy applied.
It performs well when 
$r_o=0.1\%$, but drops significantly when $r_o$ reaches 0.5\%.
That's because for a mini-batch size of 200 that we use, an outlier ratio of $0.5\%$ in training data results in an average of one outlier in each mini-batch, which 
could be learned by the baseline model.
\highlight{Note that the clipping bound $C$=$1$ also restricts the contribution of outliers in SGD steps. We conduct an ablation study which only clips the per-example gradients with $C$ without adding any noise in each gradient descent step. The results are shown as $\sigma=0$ in Table \ref{table:train_ae}. As an intermediate step to bound the sensitivity in differential privacy, clipping itself is able to slightly improve the anomaly detection results in most cases.
Still, we show that increasing the noise scale could further improve the utility.}
\revise{We highlight one of the best results in each column, and find that the trend follows our analysis in Theorem~\ref{thm:gap}. Specifically, the more outliers 
in the training dataset, the larger noise scale 
is needed for the best improvement.
As explained for (\ref{eqn:bound}), our theory shows that the privacy parameters cannot be too large or too small to ensure optimal anomaly detection performance, which coincides with the experimental results in Table~\ref{table:train_ae}.}
Although it could be challenging to select the desired noise level for training, we note that as shown in Table~\ref{table:train_ae}, applying differential privacy effectively improves the anomaly detection performance in most cases, except when $\sigma$ is too big to ruin the model parameters completely (e.g., $\sigma$=$50$) . 
Therefore,
it is generally safe and almost always helpful to apply a small amount of differential privacy noise for anomaly detection.
The noise scale could be increased further as long as the model converges.
However, it should be noted that 
applying differential privacy makes the model training much slower than the baseline.
In our experience utilizing NVIDIA Tesla V100 SXM2 GPU cards, the training time for each epoch could be up to 80 times longer.
Finally, a training data portion as high as $10\%$ might not be ``outliers'', 
but could be part of the input pattern that should be learned by the model. We show in this case, a relatively large noise scale could effectively improve the anomaly detection results (e.g., $\sigma$=$10$), but it's up to the requirement of the application whether to apply this.
\subsection{Hadoop file system log anomaly detection with LSTM}
\label{sec:exp-log}
In this section, we use a real-world example for Hadoop file system log anomaly detection, to show how anomaly detection with differential privacy outperforms the current state-of-the-art results.

\paragraph{Dataset}
The Hadoop file system (HDFS) log dataset (\cite{hdfslog}) is generated through running Hadoop map-reduce jobs for 48 hours on 203 Amazon EC2 nodes.
This dataset contains over 11 million log entries, which could be further grouped into $575,059$ block sessions by the block identifier each log has.
Each block is associated with a 
normal/abnormal label provided by domain experts.
Over the past decade this log dataset has been extensively used for research in system log 
anomaly detection (\cite{xu2009detecting, lou2010mining, du2017deeplog}).
The state-of-the-art 
results are achieved by DeepLog (\cite{du2017deeplog}), which we use as the baseline model. 
As in DeepLog, our training dataset contains $4,855$ normal block sessions, while the test dataset includes $553,366$ normal sessions and $16,838$ abnormal sessions.



\paragraph{Baseline model and metrics}
DeepLog
utilizes LSTM neural networks to learn system log sequences.
Note that system log messages are textual logs, e.g., \textit{``Transaction A finished on server B.''}. Before applying LSTM, 
a log parsing step first maps each log message into its corresponding log printing statement in the source code, e.g., \textit{``print('Transaction \%s finished on server \%s.''\%(x,y))"}. Since there are only a constant number (e.g., $N$) of log printing statements in the source code, each one could be mapped to a discrete value from a fixed vocabulary set (e.g., having size $N$).
With that, a block session of log messages could be parsed to a sequence of discrete values, e.g. \textit{``22 5 5 5 11 9 11 9 11 9 26 26 26''}.
Leveraging the fact that hidden execution paths written in source code restrict the possibilities of how one system log follows another,
DeepLog trains an LSTM model
on normal discrete sequences, which learns to predict the next discrete value given its history. In detection, the LSTM model predicts a probability distribution on all possible values that may appear at a given time step. The real executed value is detected as abnormal if it's \textit{unlikely} to happen based on LSTM prediction.
The criteria presented in DeepLog is to first sort the predicted values based on the assigned probabilities, e.g., for 
a prediction ``\textit{\{5: 0.2, 9: 0.08, 11: 0.01, 26: 0.7, ...\}}'', the order would be \textit{26, 5, 9, 11, ...}. The given value to detect is checked against the sorted top $k$ predictions, and is detected as abnormal if it's not one of them.
For anomaly detection metrics, we want to highlight that applying differential privacy significantly reduces false negatives, 
without introducing many false positives.
Therefore, we'll focus on the comparison over the number of false positives and false negatives, while also presenting measurements that indicate the overall detection performance. 

\paragraph{Set up} 
For the baseline model DeepLog, we train an LSTM model for 100 epochs, and use the final model as the anomaly detection model. The model related parameters are: 2 layers, 256 units per layer, 10 time steps, and a batch size of 256.
We call the DeepLog model with differential privacy as \textit{DeepLog+DP}.
For differential privacy, we use a clipping bound $C=1$, $\delta=10^{-5}$, and vary the noise scale $\sigma$. 
All other model related settings for DeepLog+DP are the same as DeepLog.


\begin{figure}
    \centering
    \begin{subfigure}[b]{0.493\textwidth}
    \
        \includegraphics[width=\textwidth,height=5cm]{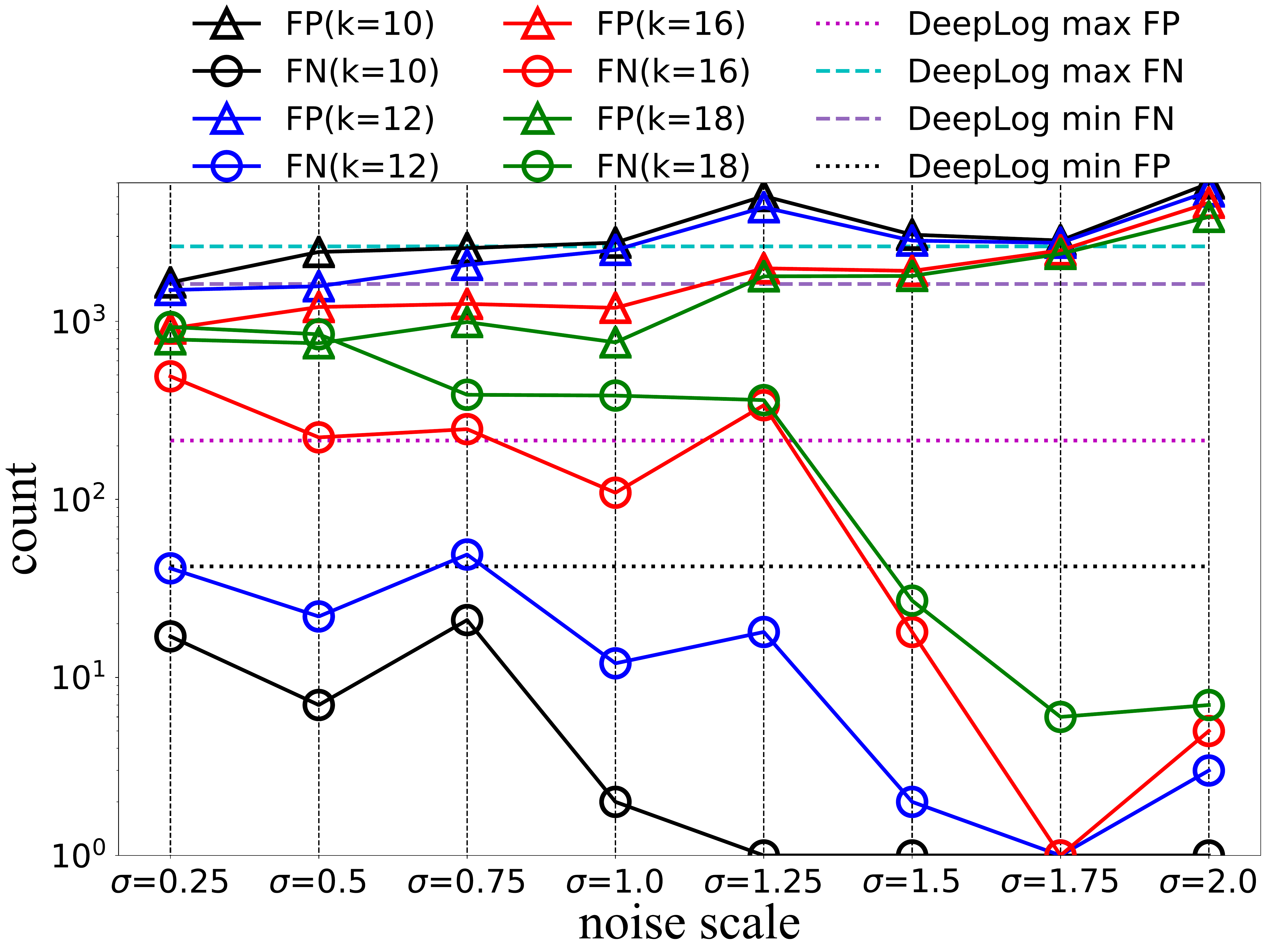}
        \caption{Comparison of \fp and \fn.}
        \label{fig:fpfn}
    \end{subfigure}
    \begin{subfigure}[b]{0.493\textwidth}
        \includegraphics[width=\textwidth,height=5cm]{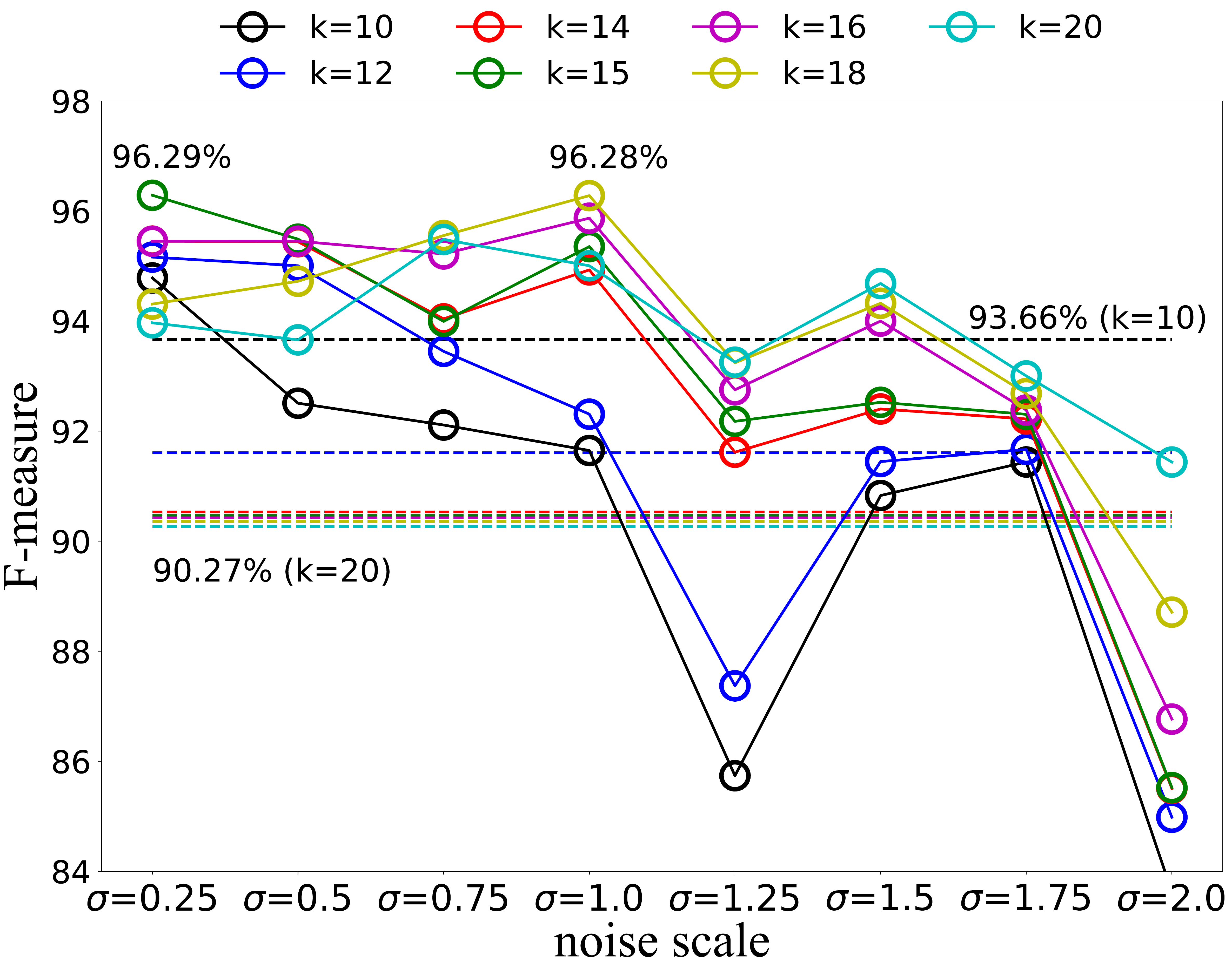}
        \caption{\fm comparison (horizontal lines: DeepLog).}
        \label{fig:fm}
    \end{subfigure}
    \caption{Improvements by differential privacy for DeepLog.}
\end{figure}

\paragraph{Results}
Figure~\ref{fig:fpfn} shows the comparison of \fp and \fn under different thresholds $k$, with the increase of noise scale $\sigma$. For clarity, we only show the following two cases for baseline model DeepLog: $k=10$ which has the maximum \fp and the minimum \fn, and $k=18$ which has the minimum \fp and the maximum \fn. Note that y axis is plotted as log scale.
It is clear that applying DP noise significantly reduces \fn in all cases, from over a thousand in DeepLog, to hundreds or even zero in DeepLog+DP. Also, the larger noise being added, the more \fn are reduced.
Although more \fp could be introduced in some cases, we note that in system anomaly detection, the merit of fewer false negatives in fact worth the cost of more false positives. 
Reported false positives could be further checked by system admin, and then fed into the model for incremental learning. However, a false negative may never be found out, until a more disastrous event occurs due to the un-discovery of it.


The \fm measurements are plotted in Figure~\ref{fig:fm}. For DeepLog model, \fm ranges from 90.38\% ($k=20$) to 93.81\% ($k=10$). 
For DeepLog+DP, the best \fm scores include 96.29\% ($\sigma=0.25$, $k=15$) and 96.28\% ($\sigma=1$, $k=18$), which show clear improvements over DeepLog model. 
Note that the best \fn and \fp measurements reported in DeepLog (\cite{du2017deeplog}) are 619 and 833 respectively, while DeepLog+DP achieves \fn=383, \fp=762 at the \fm of 96.28\% ($\sigma=1$, $k=18$); and \fn=123, \fp=1040 at the \fm of 96.29\% ($\sigma=0.25$, $k=15$), showing its ability to significantly reduce false negatives without introducing many false positives.
As shown in the figure, DeepLog performs better when $k$ is smaller, while DeepLog+DP benefits from larger $k$s. This scenario could also be explained by the addition of differential privacy noise. Since the trained model does not overfit to outliers, it assigns to anomalies much lower probabilities, 
so that anomalies are
ranked much lower than that in the DeepLog model. Meanwhile, normal execution logs are also possibly predicted with lower probabilities because of the uncertainty brought by the noise. As a result, the ideal threshold $k$ for DeepLog+DP is higher than that of DeepLog. We also note that a large noise scale could hurt the overall performance, as shown by the downward trend when $\sigma$ increases from $1.75$ to $2.0$.



\highlight{
\subsection{Backdoor attack detection}
\label{sec:exp-backdoor}
Since poisoning examples for backdoor attacks are essentially ``outlier'' training samples injected by attackers, this section conducts proof-of-concept experiments to examine whether measuring model loss as for outliers works to detect poisoning examples, and whether differential privacy is able to further improve the performance.
\revise{This detection scenario is particularly useful for backdoor attacks injected in the crowdsourcing scenario, where the model trainer gathers training data from untrusted individuals. In this case, the model trainer does not have control over the data quality but does have control over the model training process. Our proposal of adding DP noise is useful for detecting backdoor attacks and training more robust models in such a scenario.
}

\paragraph{Dataset and set up}
MNIST dataset as described in Section~\ref{sec:exp-auto} is used in this set of experiments. We refer the original 60,000 training images as \textit{CLEAN-train} and the 10,000 test images as \textit{CLEAN-test}.
We construct the backdoor attacks as described in \cite{gu2017badnets}, Section 3.1.2. Specifically, each poisoning example is generated by reversing 4 pixel values in the bottom right corner of a clean image having label $i$, and assigning backdoor label $(i+1)\%10$. From {CLEAN-train}, we randomly sample a poisoning ratio of $r_p$ images to be poisoning examples, resulting in a poisoned training dataset \textit{POISONED-train($r_p$)}.
To demonstrate the effectiveness of the poisoning attacks, we use the {CLEAN-test} dataset, as well as \textit{POISONED-test} dataset which is constructed by poisoning \textit{all} images in CLEAN-test.
For image classification model, we use convolutional neural network (CNN) containing 2 convolutional layers with max pooling, and a softmax layer to output desired labels. The differentially private models are trained 
with the same configurations
as in Section~\ref{sec:exp-auto} \revise{unless otherwise noted}.

\paragraph{Metrics}
We first evaluate the effectiveness of the constructed backdoor attack.
A successful backdoor attack should have high image classification accuracy on {CLEAN-test}, which we refer to as \textit{benign accuracy}, as well as high accuracy on {POISONED-test} with poisoned labels, which indicates the \textit{success rate}.
To investigate whether measuring the classification model loss is able to detect poisoning examples, and whether differential privacy is able to improve the detection performance, we leverage metrics AUPR and AUROC as described at the beginning of Section~\ref{sec:exp}.

\begin{table}[htbp]
    \centering
\small
    \begin{tabular}{ |l||c c||c c||c c| } \hline
    noise   &  \multicolumn{6}{|c|}{detection (AUPR / AUROC) and attack (benign accuracy / success rate) performance }    \\  
    scale    & \multicolumn{2}{|c||}{$r_p=$$0.5\%$} & \multicolumn{2}{|c||}{$r_p=$$1 \%$} & 
    \multicolumn{2}{|c|}{$r_p=$$5 \%$}     \\ \hline
     $\sigma=$  & detection & attack & detection & attack &	detection & attack      \\ \hline
     N/A  & 73.01 / 99.26 & {\bf 98.93 / 47.85} & 27.02 / 95.23 & {\bf 98.95 / 97.12} &	14.85 / 78.88 & {\bf 99.11 / 98.1}      \\ \hline
    ${0}$  &  
    91.22 / 99.92 & 97.66 / 0.23 &	92.11 / 99.88 & 97.84 / 0.35 &	95.33 / 99.79 & 97.46 / 0.3
 \\ \hline
    $  0.005$  & {\bf 92.64 / 99.9} & 97.57 / 0.17 & 94.04 / 99.93 & 97.46 / 0.28 &	94.76 / 99.79 & 97.75 / 0.3
 \\ \hline
    $0.01$ & 92.24 / 99.92 & 97.51 / 0.25 &	94.03 / 99.92 & 97.4 / 0.34 &	93.4 / 99.74 & 97.55 / 0.31 \\ \hline
    $  0.05$  &  90.76 / 99.9 & 97.42 / 0.24 & {\bf 95.11 / 99.94} & 97.8 / 0.37 &	95.09 / 99.83 & 97.72 / 0.3 \\ \hline
    $  0.1$  & 92.16 / 99.93 & 97.55 / 0.25 &	94.85 / 99.93 & 97.7 / 0.28 &	{\bf 95.33 / 99.82} & 97.34 / 0.39
 \\ \hline
    \end{tabular}
	\caption{{Backdoor attack and detection results with varying poisoning ratio $r_p$ (clipping bound $C=1$).} 
	}
	\label{table:train_backdoor}
\end{table}

\paragraph{Results}
\revise{We first evaluate the backdoor attack effectiveness and the detection performance with varying poisoning ratio $r_p$, under different noise scale $\sigma$, and fixed clipping bound $C=1$.} The results are summarized in Table~\ref{table:train_backdoor}. $\sigma=$N/A indicates classification models trained without differential privacy. Benign accuracy remains high on clean data. Backdoor success rate is only around half at a poisoning ratio of 0.5\%, and shows successful (97.12\% success rate) at a poisoning ratio of 1\%.
Detecting poisoning examples by measuring model loss shows some level of effectiveness when the poisoning ratio is low (e.g., 0.5\%). Furthermore, applying differential privacy to the model training process is able to significantly improve the detection performance. Similar as in Table~\ref{table:train_ae}, the higher the poisoning ratio, the larger the noise level (smaller $\epsilon$) to achieve the best improvement.
Another interesting observation is that, a differentially private model is naturally robust to backdoor attacks. As indicated in Table~\ref{table:train_backdoor}, differential privacy effectively limits the success of backdoor attacks, reducing the success rate below 0.5\% in \textit{all} cases. In comparison, the utility downgrade on benign accuracy is little.

\revise{To further evaluate the applicability of using the same CNN model for both anomaly detection and image classification,
seeking to co-optimize the performance of the model for both tasks, we collect measurements with varying clipping bound $C$, and fixed noise scale $\sigma=0.5$, as a complement to Table~\ref{table:train_backdoor}. The results are summarized in Table~\ref{table:train_backdoor_c}. Note that a small $C$ may hurt model performance as more model parameters are clipped. When $C$ is greater than most parameter values, the effect of increasing $C$ is similar to that of increasing $\sigma$~\cite{abadi2016deep}.
From Table~\ref{table:train_backdoor_c}, we can observe that the best model for anomaly detection could have a similar set of parameters with the best model for image classification. 
However, in general, as shown in Table~\ref{table:train_backdoor}, classification accuracy and robustness are often two conflicting desiderata; model trainers can tune the privacy parameter in order to meet the task-specific requirements for accuracy and robustness.
} 

\begin{table}[htbp]
    \centering
\small
    \begin{tabular}{ |l||c c||c c||c c| } \hline
    clipping   &  \multicolumn{6}{|c|}{detection (AUPR / AUROC) and attack (benign accuracy / success rate) performance }    \\  
    bound    & \multicolumn{2}{|c||}{$r_p=$ $0.5\%$} & \multicolumn{2}{|c||}{$r_p=$ $1 \%$} & 
    \multicolumn{2}{|c|}{$r_p=$ $5 \%$}     \\ \hline
     $C=$  & detection & attack & detection & attack &	detection & attack      \\ \hline
0.5 &	87.46 / 99.87 &	96.29 / 0.31 & 90.78 / 99.85 &	96.47 / 0.26 & 95.62 / 99.79 & 96.73 / 0.34 \\ \hline
0.8 & 89.03 / 99.85 & 97.13 / 0.33 & 92.39 / 99.89 & 97.11 / 0.32 & {\bf 95.63 / 99.79} & {\bf 97.4} / 0.28 \\ \hline
1 & 90 / 99.9 & 97.28 / 0.22 &  93.46 / 99.92 &  {\bf 97.47} / 0.25 & 95.37 / 99.79 & 97.34 / 0.3 \\ \hline
2 & {\bf 90.85 / 99.81} &  {\bf 97.48} / 0.24 & {\bf 93.49 / 99.91} & 97.21 / 0.26 & 93.26 / 99.75 & 97.39 / 0.46 \\ \hline
3 & 90.17 / 99.93 & 97.29 / 0.3 & 88.05 / 99.84 & 97.18 / 0.33 & 89.51 / 99.59 & 97.35 / 0.48 \\ \hline
    \end{tabular}
	\caption{Backdoor attack and detection results with  varying poisoning ratio $r_p$ (noise scale $\sigma=0.5$).  
	}
	\label{table:train_backdoor_c}
\end{table}

}

\vspace{-2mm}
\section{Related work\todo{related work for backdoor detection?}}
\vspace{-2mm}
To the best of our knowledge, this paper is the first one that proposes to improve outlier/novelty detection with differential privacy, and further extends it to backdoor attack detection. 
\highlight{Note that this is not the first work that combines outlier detection and differential privacy together. \cite{okada2015differentially} aim to 
preserve input data privacy while detecting outliers. The two tasks are contradicting in this case as the identification of outliers (part of input data) implies certain privacy leakage, so \cite{okada2015differentially} try to find a balance. In contrast, we focus on improving anomaly detection performance with differential privacy, which is only applied to the model training stage, but no privacy protection is provided for the input data in detection stage when the outliers are actually being detected.}

Outlier detection and novelty detection
are closely related to each other and often addressed together (\cite{hodge2004survey, sk-outlier-novelty, scikit-learn}).
Outlier detection is the process of identifying rare items in a dataset that significantly differ from the majority (\cite{aggarwal2001outlier}), while novelty detection is to detect new observations that lie in the low density area of the existing dataset (\cite{markou2003novelty1, markou2003novelty2}).
Previous work mostly achieves outlier detection using unsupervised learning methods (\cite{zimek2012survey, zimek2014ensembles, campos2016evaluation}), 
while novelty detection typically assumes a normal dataset is available for training, and is realized by semi-supervised learning (\cite{blanchard2010semi, de2013semi}). In both cases, it involves summarizing a distribution that the majority of training data are drawn from.
Traditional methods such as clustering (\cite{duan2009cluster}) and principal component analysis (PCA) (\cite{xu2010robust, hoffmann2007kernel}) have been frequently used. 
In this paper, 
we leverage deep learning based detection methods including autoencoders (\cite{gottschlich2017autoperf}) and LSTM (\cite{du2017deeplog}) as the baselines, and further extend the idea of measuring model loss to backdoor attack detection.

Proposed by \cite{differential}, differential privacy has been a powerful tool to protect input data privacy.
\cite{kasiviswanathan2011can} shows that differential privacy implies stability on the output statistical results.
Further, \cite{dwork2015preserving}  points out 
that the empirical average of the output of a differentially private algorithm on a random dataset is close to the true expectation with high probability.
Differential privacy has been utilized to train machine learning models that are robust to adversarial examples (\cite{phan2019preserving, lecuyer2018certified}), and to bound the success of 
inference attacks (\cite{yeom2018privacy}).
In this paper, we utilize the property of differential privacy to improve anomaly detection and privacy is ensured via the technique proposed in~\cite{abadi2016deep}.

\revise{Lastly, we note that a recent paper by \cite{bagdasaryan2019differential} showed that accuracy of
differentially private models drops much more for the underrepresented classes and subgroups. Intrinsically, our paper exploited the same phenomenon to improve anomaly detection. \cite{bagdasaryan2019differential} studied the phenomenon empirically, while our work provides a theoretical analysis, which, for the first time, precisely characterizes the dependence of the performance gap between the majority and the underrepresented group on the privacy parameters. Moreover, \cite{bagdasaryan2019differential} mainly considered the implication of differential privacy to the fairness of machine learning models; by contrast, our paper focuses on anomaly detection and backdoor attacks and exhibits strong empirical evidence for the efficacy of differential privacy in these two application domains.}

\vspace{-2mm}
\section{Conclusion}
\vspace{-2mm}
In this paper, \revise{inspired by the fact that differential privacy implies stability, we apply DP noise to 
improve the performance of outlier detection and novelty detection, with an extension to backdoor attack detection. We first provide the theoretical basis for the efficacy of differential privacy for identifying anomalies, connecting the hardness of the identification problem to privacy parameters. Our theoretical results are useful to explain various experimental findings, including how the anomaly detection performance varies with privacy parameters and the number of outliers in the training set. We perform extensive experiments to demonstrate the effectiveness of differential privacy for anomaly detection. To fully evaluate the effectiveness of DP in anomaly detection with different amount of outliers and noisee, we first construct a contaminated dataset based on MNIST and train autoencoder anomaly detection models with varying noise scale applied.}
We then evaluate the performance using a real-world task, Hadoop file system log anomaly detection, by applying DP noise to DeepLog, the current state-of-the-art detection model. 
The evaluation results show that DP noise is effective towards reducing the number of false negatives, and further improving the overall utility. Finally, we generalize the idea of measuring model loss for outlier detection to backdoor attack detection \revise{ and further improve the performance via differential privacy.}


\bibliographystyle{unsrt}  
\bibliography{main_arxiv}  

\end{document}